\documentclass{article}

 \usepackage[dblblindworkshop, final]{neurips_2025}

\workshoptitle{Foundation Models for the Brain and Body}

\usepackage[utf8]{inputenc} 
\usepackage[T1]{fontenc}    
\usepackage{hyperref}       
\usepackage{url}            
\usepackage{booktabs}       
\usepackage{amsfonts}       
\usepackage{nicefrac}       
\usepackage{microtype}      
\usepackage{xcolor}   
\usepackage{graphicx}
\usepackage{float}
\usepackage{amsmath}
\hypersetup{
    pdfborder = {0 0 0} 
}

\title{Leveraging Foundational Models and Simple Fusion for Multi-modal Physiological Signal Analysis}

\author{%
  Youssef Ghallab\textsuperscript{\textnormal{1}\textnormal{,2} }\hspace{0.1cm} 
  Omar Iraqy\textsuperscript{\textnormal{1}} \hspace{0.1cm}
  Mohamed Kandil\textsuperscript{\textnormal{1}} \hspace{0.1cm}
  Mohamed Ashraf\textsuperscript{\textnormal{1}} \\
  \And
  Saadeldine Eletter\textsuperscript{\textnormal{1}\textnormal{,2}}\hspace{0.1cm}   
  Morougue Ghazal\textsuperscript{\textnormal{1}} \hspace{0.1cm}
  Ayman Khalafallah\textsuperscript{\textnormal{1}} \hspace{0.1cm}
  Nagwa El-Makky\textsuperscript{\textnormal{1}}\\
  \\
  \textsuperscript{1}Computer and Communication Engineering Department, Alexandria University \\
    \textsuperscript{2}Mohamed bin Zayed University of Artificial Intelligence\\
  \and
  \{es-youssifghalab2025, es-omarmohammed2025, es-mohamed.abdelmoneim2025,\\ es-mohameda.hamdy2025, es-saadeddin2025, es-morojmahmoud2025, ayman.khalafallah, \\nagwamakky\}@alexu.edu.eg\\
  \and 
  \{youssef.ghallab, saadeldine.eletter\}@mbzuai.ac.ae
}

\begin{document}

\maketitle

\begin{abstract}
    Physiological signals such as electrocardiograms (ECG) and electroencephalograms (EEG) provide complementary insights into human health and cognition, yet multi-modal integration is challenging due to limited multi-modal labeled data, and modality-specific differences . In this work, we adapt the CBraMod encoder~\cite{cbramod} for large-scale self-supervised ECG pretraining, introducing a dual-masking strategy to capture intra- and inter-lead dependencies. To overcome the above challenges, we utilize a pre-trained CBraMod encoder~\cite{cbramod} for EEG and pre-train a symmetric ECG encoder, equipping each modality with a rich foundational representation. These representations are then fused via simple embedding concatenation, allowing the classification head to learn cross-modal interactions, together enabling effective downstream learning despite limited multi-modal supervision. Evaluated on emotion recognition, our approach achieves near state-of-the-art performance, demonstrating that carefully designed physiological encoders, even with straightforward fusion, substantially improve downstream performance. These results highlight the potential of foundation-model approaches to harness the holistic nature of physiological signals, enabling scalable, label-efficient, and generalizable solutions for healthcare and affective computing.

\end{abstract}

\section{Introduction}

Physiological signals represent a rich source of information about human health, cognitive states, and emotional responses. Among these, electrocardiograms (ECG) and electroencephalograms (EEG) offer complementary perspectives on human physiology—ECG capturing cardiac dynamics that reflect autonomic nervous system activity, while EEG provides direct measurements of neural oscillations underlying cognitive and emotional processes.

The integration of these modalities holds significant promise for advancing healthcare diagnostics, brain-computer interfaces, and affective computing applications. Despite this potential, multi-modal physiological signal analysis faces several fundamental challenges. First, the scarcity of multi-modal labeled datasets limits the development of supervised learning approaches, as simultaneous recording of multiple physiological signals with expert annotations remains expensive and time-intensive. Second, the inherent differences in signal characteristics—ECG's relatively regular morphology versus EEG's complex spatiotemporal patterns—create modality-specific representational gaps that complicate joint analysis. Third, traditional approaches often rely on hand-crafted features and task-specific architectures, limiting their generalizability across diverse applications and patient populations.

Recent advances in foundation models have demonstrated remarkable success in learning universal representations across domains such as natural language processing and computer vision \cite{bert}. These models leverage self-supervised pretraining on large-scale unlabeled data to capture generalizable patterns that can be adapted to downstream tasks with minimal supervision. In the physiological signal domain, this paradigm offers particular promise given the abundance of unlabeled biosignal data and the potential for transferable representations that capture fundamental physiological processes.

However, adapting foundation model approaches to physiological signals presents unique challenges. Unlike text or images, physiological signals exhibit complex temporal dependencies, multi-scale patterns, and inherent variability across subjects and recording conditions. Furthermore, the effective integration of multiple physiological modalities requires careful consideration of how to align and fuse representations that capture both modality-specific characteristics and cross-modal relationships.

In this work, we address these challenges through a foundation model approach that combines self-supervised pretraining with strategic multi-modal fusion. We adapt the CBraMod architecture \cite{cbramod}, originally designed for EEG analysis, to create a symmetric ECG encoder capable of learning robust cardiac representations. Our ECG pretraining framework introduces a dual-masking strategy that encourages the model to capture both intra-lead temporal patterns and inter-lead spatial dependencies, enabling comprehensive understanding of cardiac dynamics.

Rather than pursuing complex multi-modal alignment schemes, we demonstrate that carefully designed modality-specific encoders combined with straightforward embedding concatenation can achieve highly effective cross-modal learning. This approach leverages the rich foundational representations learned by each encoder while allowing the downstream classification head to discover cross-modal interactions relevant to the target task.
We evaluate our framework on emotion recognition using the DREAMER dataset \cite{Dreamer}, where our approach achieves near state-of-the-art performance across valence, arousal, and dominance dimensions. Our results demonstrate that foundation model approaches can effectively harness the complementary nature of physiological signals, providing a scalable and label-efficient pathway for multi-modal biosignal analysis.

\section{Method}
\label{methodology}

\begin{figure}[H]
  \centering
  \includegraphics[width= \textwidth]{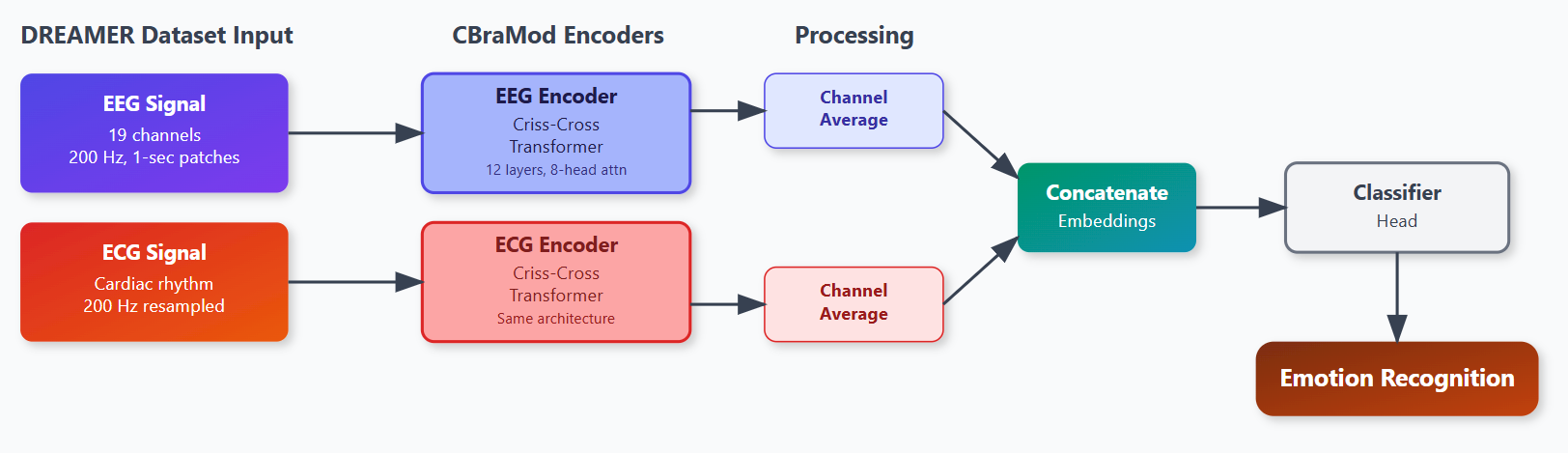}
  \caption{Architecture of the proposed multimodal emotion recognition framework. Electroencephalogram (EEG) and electrocardiogram (ECG) signals are processed by separate encoders to extract modality-specific features. The resulting embeddings are averaged, concatenated, and passed to a classifier for emotion prediction.}
  \label{architecture}
\end{figure}

Figure~\ref{architecture} provides an overview of our multi-modal framework for emotion recognition using ECG and EEG signals from the DREAMER dataset [2]. EEG and ECG signals are first processed by separate respective encoders with similar architectures, capturing modality-specific features. After encoding, a channel-wise averaging step reduces each representation to a compact embedding. The embeddings from both modalities are then concatenated and passed through a lightweight classifier head, which learns cross-modal interactions to perform the downstream emotion recognition task.

\subsection{ECG Pre-training}

To address the scarcity of large-scale multimodal datasets, we pre-train an encoder tailored for ECG signals, ensuring it can provide rich and transferable representations.

\paragraph{Architecture} We adopt the CBraMod~\cite{cbramod} architecture, specifically a criss-cross transformer with 12 layers and 8 attention heads, where input ECG patches are first embedded and enriched with positional encodings. The criss-cross attention mechanism enables the model to jointly capture local temporal dynamics and global contextual dependencies. This design is kept identical to the EEG encoder, ensuring architectural symmetry across modalities and facilitating comparable representation spaces for downstream fusion. 

\paragraph{Pre-processing} For ECG pre-training, we employ a multi-stage preprocessing pipeline to ensure signal quality and robustness across acquisition settings. First, a Butterworth bandpass filter (0.5–40 Hz) is applied to isolate clinically relevant ECG components while suppressing baseline wander and high-frequency noise, followed by a notch filter (50 Hz or 60 Hz, depending on region) to attenuate powerline interference. To address variability in acquisition hardware, we adopt a multi-rate resampling strategy, standardizing signals to 100, 200, 500, and 1000 Hz, which exposes the model to diverse sampling resolutions and promotes invariant feature learning. Each ECG is then segmented into fixed 200-sample windows and reshaped into tensors of size (12, 10, 200), corresponding to 12 clinical leads, 10 temporal patches, and 200 samples per patch, with leads reordered to follow the conventional clinical configuration.

\paragraph{ECG Patching \& Dual-masking}

We partition each ECG sample \(S \in \mathbb{R}^{C \times T}\), where \(C\) denotes the number of leads and \(T\) the number of timepoints, into \(n = \lfloor T/t \rfloor\) temporal patches per channel, resulting in the patch set:

\begin{equation}
X = \{x_{i,j} \mid i \in [1,\dots,C],\; j \in [1,\dots,n]\}.    
\end{equation} 

To enhance robustness across both spatial (lead-wise) and temporal contexts, we apply a \emph{dual-masking} scheme:
\begin{itemize}
  \item \textbf{Patch masking}: a binary mask \(M \in \{0,1\}^{C \times n}\) is generated, with each \(m_{i,j}\) sampled from a Bernoulli distribution with masking ratio \(r\); if \(m_{i,j} = 1\), the corresponding patch \(x_{i,j}\) is masked.
  \item \textbf{Channel masking}: independently, we select a subset of channels \(\mathcal{C}_m \subseteq \{1, \dots, C\}\) using masking proportion \(r_c\); all patches in these channels are masked to encourage learning across lead configurations.
\end{itemize}
The resulting masked patch \(\tilde x_{i,j}\) is defined as:
\begin{equation}
\tilde x_{i,j} =
\begin{cases}
x_M, & \text{if } m_{i,j} = 1 \text{ or } i \in \mathcal{C}_m, \\
x_{i,j}, & \text{otherwise},
\end{cases}
\end{equation}
and the full masked input becomes:
\begin{equation}
\tilde X = \{\tilde x_{i,j} \mid i \in [1,\dots,C],\; j \in [1,\dots,n]\}.
\end{equation}
This dual-masking encourages the encoder to learn both local temporal patterns and global inter-lead dependencies, improving robustness to channel-specific artifacts and missing data.

\paragraph{Learning Objective} 

The learning objective in our ECG pre-training is designed to capture both fine-grained temporal dynamics and inter-lead dependencies by employing a composite mean squared error (MSE) loss. Similar to CBraMod~\cite{cbramod}, a reconstruction head is used for reconstructing the masked patches and channels. 

We use the same patch reconstruction loss as in CBraMod and add a new channel reconstruction loss, which ensures the model can recover signals from entirely masked leads. we compute the reconstruction loss for channels that were masked entirely and those that were not:

\begin{equation}
\mathcal{L}_{\text{channel}} = \frac{1}{2} \left( \| \hat{X}_M^c - X_M^c \|_2^2 + \| \hat{X}_{\bar{M}}^c - X_{\bar{M}}^c \|_2^2 \right)
\end{equation}

\noindent where $X_M^c$ and $\hat{X}_M^c$ represent the masked channel data and its reconstruction, and $X_{\bar{M}}^c$, $\hat{X}_{\bar{M}}^c$ denote the unmasked channels.

To promote balanced learning, the loss is computed over both masked and inverse-masked regions, assigning equal weight to each.

The total loss is an equally weighted summation of both components:

\begin{equation}
\mathcal{L}_{\text{total}} = \frac{1}{2} \mathcal{L}_{\text{patch}} + \frac{1}{2} \mathcal{L}_{\text{channel}}
\end{equation}

This loss design ensures the model learns to reconstruct both localized temporal content and complete signals across leads, leading to generalized and transferable ECG representations.

\subsection{Multi-modal Fusion Framework}

To leverage complementary information from ECG and EEG signals, we adopt a dual-backbone approach, employing our pre-trained ECG foundational model as the ECG encoder and the CBraMod model~\cite{cbramod} as the EEG encoder. The embeddings produced by each backbone are then fused through straightforward concatenation, which is subsequently passed to a feed-forward classification head.

Let:
\begin{equation}    
\mathbf{E}_{\text{ECG}} = f_{\text{ECG}}(\mathbf{X}_{\text{ECG}})
\end{equation}

be the embedding produced by the pre-trained ECG backbone for an input ECG signal $\mathbf{X}_{\text{ECG}}$, and
\begin{equation}    
\mathbf{E}_{\text{EEG}} = f_{\text{EEG}}(\mathbf{X}_{\text{EEG}})
\end{equation}
be the embedding produced by the pre-trained CBraMod EEG backbone for an input EEG signal $\mathbf{X}_{\text{EEG}}$.

Fusion via concatenation:
\begin{equation}    
\mathbf{E}_{\text{fusion}} = \text{Concat}(\mathbf{E}_{\text{ECG}}, \mathbf{E}_{\text{EEG}})
\end{equation}

Classification head:
\begin{equation}
\hat{\mathbf{y}} = g(\mathbf{E}_{\text{fusion}}) = g(\text{Concat}(f_{\text{ECG}}(\mathbf{X}_{\text{ECG}}), f_{\text{EEG}}(\mathbf{X}_{\text{EEG}})))
\end{equation}

where $g(\cdot)$ denotes a feed-forward network that maps the fused embedding to the task-specific prediction $\hat{\mathbf{y}}$.

This framework enables the classification network to learn a unified representation that captures cross-modal interactions, making it well-suited for downstream tasks that require multi-modal information while preserving meaningful modality-specific features through the respective encoders.

\section{Experiments \& Results}

\subsection{ECG Pretraining Setup}

We utilize the PhysioNet/Computing in Cardiology Challenge 2021 dataset~\cite{physionet2021}, which is a collection of seven distinct datasets, for ECG pretraining, comprising 88,253 12-lead ECG recordings from diverse geographic and clinical sources across three continents.. This multi-dataset composition introduces substantial variability in patient demographics, healthcare settings, and device configurations, with recordings varying in length (up to 60 seconds) and sampling rates (500–1000 Hz). During self-supervised pretraining, diagnostic labels are omitted to learn task-agnostic representations purely from signal structure and temporal-spatial patterns. All pretraining experiments are conducted on dual NVIDIA A100 GPUs.

\subsection{Evaluation on Emotion Recognition}

We evaluate our approach on the DREAMER dataset~\cite{Dreamer}, which contains EEG and ECG recordings from 23 subjects watching emotional video stimuli. Each subject viewed 18 video clips while continuous EEG (14 channels at 128 Hz) and ECG (2 channels at 256 Hz) were recorded. Emotional responses were self-assessed on continuous scales for valence, arousal, and dominance, which we discretize into binary classification tasks (Low: $<3$, High: $\geq 3$).

Our model processes both modalities through independent CBraMod encoders, where the EEG encoder uses pretrained weights from the Temple University Hospital EEG Corpus (TUEG) and the ECG encoder uses our self-supervised pretraining weights. Embeddings are averaged across channels, concatenated, and passed to three parallel binary classification heads. We employ subject-independent 3:1:1 train/validation/test splitting and train using Adam optimizer with learning rate $10^{-3}$ for 10 epochs.

\subsection{Results}

Table~\ref{tab:main_results} compares our approach against state-of-the-art methods on the DREAMER dataset. Our pretrained model achieves competitive performance across all emotional dimensions, with particularly strong results in AUC metrics for arousal (84.79) and dominance (86.69), ranking best and second-best respectively. For valence prediction, we achieve second-best accuracy (69.44\%) and best F1-score (81.14), demonstrating strong discriminative power while approaching the performance of Brant-X~\cite{brantx} with a more efficient architecture.

\begin{table*}[h]
\centering
\caption{Performance comparison on DREAMER dataset. Values show mean across subjects. \textbf{Bold} = best, \underline{Underline} = second-best.}
\label{tab:main_results}
\resizebox{\textwidth}{!}{
\begin{tabular}{l|ccc|ccc|ccc}
\toprule
\textbf{Method} & \multicolumn{3}{c|}{\textbf{Valence}} & \multicolumn{3}{c|}{\textbf{Arousal}} & \multicolumn{3}{c}{\textbf{Dominance}} \\
\cline{2-10}
 & Acc. & F1 & AUC & Acc. & F1 & AUC & Acc. & F1 & AUC \\
\midrule
TF-C~\cite{Tf-C} & 66.20 & 78.09 & 69.71 & 76.45 & 85.86 & 80.40 & 78.17 & 87.01 & 85.20 \\
SimMTM~\cite{SimMTM} & 63.84 & 75.52 & 69.73 & 76.16 & 86.21 & 76.42 & 78.54 & 87.81 & 82.84 \\
OneFitsAll~\cite{OneFitsAll} & 63.51 & 76.12 & 68.26 & 73.88 & 84.15 & 78.64 & 77.41 & 86.92 & 85.59 \\
Time-LLM~\cite{Time-LLM} & 68.03 & 72.22 & \textbf{80.83} & 76.39 & 85.63 & 80.00 & 78.47 & 85.92 & 79.68 \\
MiniRocket~\cite{MiniRocket} & 60.54 & 65.68 & 64.36 & 75.73 & 85.75 & 77.90 & 75.11 & 85.28 & \underline{86.69} \\
MLF-CapsNet~\cite{MLF-CapsNet} & 65.67 & 77.06 & 71.05 & 74.56 & 84.98 & 79.80 & 77.13 & 86.94 & 82.61 \\
EEGConformer~\cite{EEGConformer} & 59.82 & 69.53 & 71.94 & 73.07 & 83.21 & 75.11 & 81.82 & 89.50 & 83.19 \\
Lin et al.~\cite{Lin-et-al.} & 66.47 & 79.50 & 67.10 & 75.54 & 85.87 & 79.06 & 78.46 & 87.83 & 79.40 \\
Wang et al.~\cite{wang-et-al.} & 66.95 & 79.84 & 66.20 & \underline{76.47} & \underline{86.44} & 80.29 & \underline{81.87} & \underline{89.96} & 83.97 \\
Brant-X~\cite{brantx} & \textbf{70.61} & \underline{80.51} & \underline{72.48} & \textbf{78.64} & \textbf{87.59} & \underline{82.14} & \textbf{83.54} & \textbf{90.97} & \textbf{90.19} \\
\midrule
\textbf{Ours} & \underline{69.44} & \textbf{81.14} & 68.69 & 75.00 & 84.21 & \textbf{84.79} & 79.63 & 88.30 & \underline{86.69} \\
\bottomrule
\end{tabular}
}
\end{table*}

\section{Conclusion}

In this work, we presented a multi-modal fusion framework that leverages foundation models for both ECG and EEG, addressing the challenges of limited labeled data and modality-specific differences. By pretraining a symmetric ECG encoder and employing the CBraMod EEG encoder, we obtained rich modality-specific representations, which were fused via simple concatenation to enable effective cross-modal learning. Our approach demonstrated near state-of-the-art performance on emotion recognition, underscoring the effectiveness of combining carefully designed physiological encoders with straightforward fusion strategies. These findings highlight the potential of foundation-model approaches to advance scalable, label-efficient, and generalizable solutions for healthcare and affective computing.

\medskip
{
\small
\bibliographystyle{plain}

}


\appendix

\section{Pre-training Additional Details}

 To evaluate the effectiveness of our self-supervised pretraining approach, we monitor the reconstruction loss—specifically, the mean squared error (MSE) between the masked input segments and the model’s reconstructed outputs. This metric directly reflects the model’s ability to learn meaningful latent representations of the ECG signal, capturing
 both temporal and inter-lead dependencies.

 \begin{figure}[H]
  \centering
  \includegraphics[width= \textwidth]{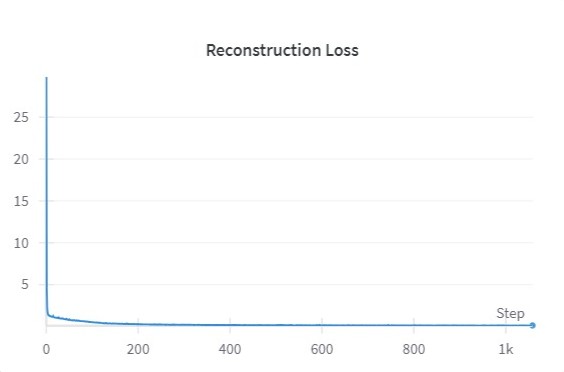}
  \caption{Reconstruction loss in pre-training}
  \label{reconstruction_loss}
\end{figure}

As shown in Figure~\ref{reconstruction_loss}, the reconstruction loss dropped significantly from an initial
 value of approximately 30 to a final value of 0.1136. This steady decline indicates that
 the model has progressively improved its understanding of the underlying ECG structures,
 even in the absence of supervised labels.

 \section{Training details for the multi-modal experiment}

This section provides details about the experimental setup and plots for the multi-modal fusion experiment.

 \begin{table}[H]
\centering
\caption{Training Hyperparameters}
\label{tab:hyperparams}
\begin{tabular}{lc}
\toprule
\textbf{Parameter} & \textbf{Value} \\
\midrule
Epochs & 10 \\
Batch size & 8 \\
Initial learning rate & $1\times10^{-3}$ \\
Classifier dropout & 0.2 \\
Optimizer & Adam \\
LR Schedular & StepLR \\ 
Gamma & 0.1 \\
Step Size & 4 epochs \\
Loss Function & BCE Loss \\ 
\bottomrule
\end{tabular}
\end{table}

\begin{figure}[H]
  \centering
  \includegraphics[width= \textwidth]{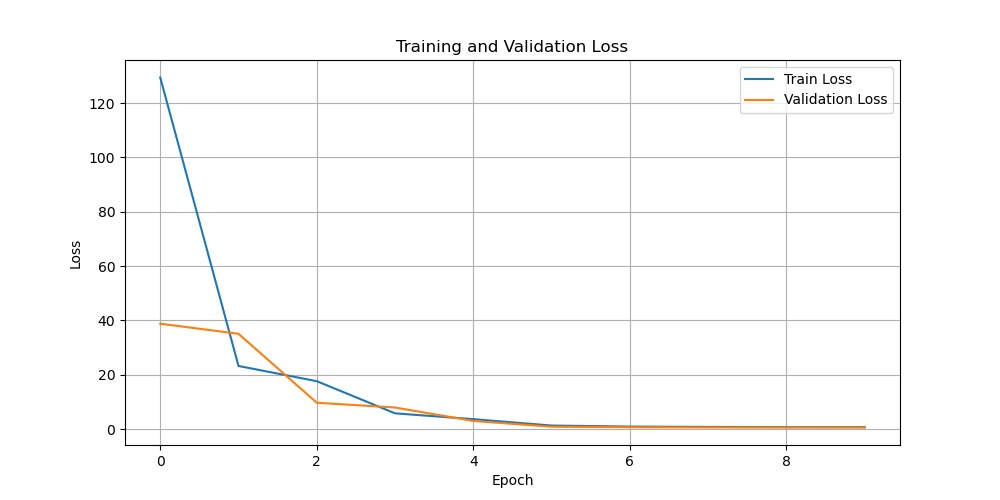}
  \caption{Training and validation loss curves over epochs for the multi-modal fusion experiment on emotion recognition task using Dreamer dataset.}
  \label{training_loss}
\end{figure}

\section{Limitations and Future Work}

Despite the promising results, our work has several limitations. First, the fusion strategy relies on simple embedding concatenation, which, while effective, may not fully exploit fine-grained temporal or cross-modal dependencies. Second, our evaluation is constrained to emotion recognition task, leaving open questions regarding generalization to broader physiological applications and more diverse populations. Future work will explore more advanced fusion mechanisms, extend evaluation to additional downstream tasks, and investigate new strategies for mitigating data scarcity, such as weak supervision or synthetic data augmentation.

\end{document}